\documentclass[manuscript]{acmart}
\AtBeginDocument{%
  }

\setcopyright{acmlicensed}
\copyrightyear{2018}
\acmYear{2018}
\acmDOI{XXXXXXX.XXXXXXX}

\acmJournal{JOCCH}
\acmVolume{37}
\acmNumber{4}
\acmArticle{111}
\acmMonth{8}

\usepackage{todonotes}

\begin{document}

\title{Multimodal Cultural Heritage Knowledge Graph Extension with Language and Vision Models}

\author{Yang Zhang}
\authornote{Work done during internship at Center for Studies and Research in Computer Science and Communication, CNAM}
\email{yang.zhang@telecom-paris.fr}
\affiliation{%
  \institution{Télécom Paris, Institut Polytechnique de Paris}
  \city{Paris}
  \country{France}
}

\author{Nada Mimouni}
\email{nada.mimouni@cnam.fr}
\affiliation{%
  \institution{Center for Studies and Research in Computer Science and Communication, CNAM }
  \city{Paris}
  \country{France}
}
\author{Jean-Claude Moissinac}
\email{jean-claude.moissinac@telecom-paris.fr}
\affiliation{%
  \institution{LTCI, Télécom Paris, Institut polytechnique de Paris}
  \city{Paris}
  \country{France}
}
\author{Fayçal Hamdi}
\email{faycal.hamdi@cnam.fr}
\affiliation{%
  \institution{Center for Studies and Research in Computer Science and Communication, CNAM}
  \city{Paris}
  \country{France}
}

\renewcommand{\shortauthors}{Zhang et al.}

\begin{abstract}
The preservation and interpretation of cultural heritage increasingly rely on digital technologies, among which Knowledge Graphs (KGs) stand out for their ability to structure vast amounts of data. However, the construction and expansion of these KGs often face challenges due to the diverse and complex nature of cultural heritage information. In this paper, we propose a novel approach for extending KG resources in the domain of cultural heritage, which we applied to French data. First, we introduce a new knowledge graph in the domain of French cultural heritage, WJoconde, which is distinguished by its multimodality as it integrates both textual and image information of the entities. We further introduce three variants of WJoconde to facilitate downstream research, such as Knowledge Graph Completion (KGC). We also built a comprehensive benchmark for KGC methods on our dataset. Second, we propose a new framework for extending cultural heritage KGs using multi-modal approaches leveraging Large Language Models (LLMs) and Vision-Language Models (VLMs), which includes automated data extraction from unstructured resources combined with a special validation pipeline for grounding the output of both models, to further extend WJoconde. Our results show that by integrating the rich text and image information in cultural heritage data, we can efficiently enhance KGs with high reliability. We open-source all code and benchmark datasets with text and images, as well as the original data with an interactive access point~\footnote{See \ref{data-availability} for data and code availability.}.
\end{abstract}



\received{20 February 2007}
\received[revised]{12 March 2009}
\received[accepted]{5 June 2009}

\maketitle

\section{Introduction}\label{sec1}

A knowledge graph (KG) represents real-world knowledge by sets of facts in the form of triples $<h,r,t>$, where the head entity (h) and tail entity (t) of a triple are two nodes of the graph, and the relation (r) is an edge connecting the two entities qualified by a predicate \cite{hogan2021kg}. KGs are valuable background resources for various applications, e.g. recommendation \cite{alghossein2018}, web search \cite{Szumlanski2010}, question answering \cite{Yin2016}, relation extraction \cite{weston-etal-2013-connecting} and RAG \cite{lewis2021retrievalaugmentedgenerationknowledgeintensivenlp}. In the domain of cultural heritage, for example, paintings have various aspects such as name, creator, time, description, and image. The management of such extensive collections has become increasingly complex over the years and has posed significant challenges for curators and researchers, especially in AI-related applications. KGs have thus become a natural and promising approach to representing cultural heritage datasets \cite{DBLP:conf/jcdl/TanTBODS21,carreiero2019-arco}, facilitating sophisticated querying and analytics, and providing a structured semantic context that is valuable for various applications. 

At the same time, international communities such as the GLAM Labs network and AI4LAM have emerged to support innovation and experimentation with digital cultural collections. GLAM Labs promotes collaborative environments where galleries, libraries, archives, and museums can explore new methods for working with digitized and born-digital heritage data \cite{DBLP:journals/corr/abs-2509-08710}, while AI4LAM \footnote{\url{https://sites.google.com/view/ai4lam/home}  (checked on 17/12/2025)} provides a framework for sharing practices and developing AI-based approaches across the LAM sector. These initiatives reflect a broader movement toward leveraging multi-modal data and AI techniques to improve access, interoperability, and knowledge extraction in cultural heritage collections.

Despite the usefulness of a structured representation, text and images, such as Wikipedia content, are rich and valuable resources for cultural heritage objects in an unstructured format \cite{agirre-etal-2012-matching, 10.1145/3012285}. We believe that incorporating such information into the creation and extension of cultural heritage KGs can provide a deeper understanding of cultural connections and historical insights, thus facilitating research and application in this domain. 

Recognizing both the power of structured KGs and the diversity of unstructured sources, this paper introduces a new knowledge graph WJoconde, focusing on the French cultural heritage; it integrates multi-modal information including text descriptions and visual images. WJoconde covers 28k French cultural heritage entities, with more than 483k triples of 370 relationships. It is built from the Joconde database \cite{moissinac_jean_claude_2020_3986498} - a French reference database on cultural heritage. WJoconde thus serves as a high-quality knowledge resource, the integration of texts and visual images further extends its usage and impact on various applications, such as querying and novel knowledge graph completion research. 

During the creation of WJoconde, we recognize  one of the main problems with knowledge graphs, even large, well-established ones
such as DBpedia \cite{DBLP:conf/semweb/AuerBKLCI07} and Freebase \cite{10.1145/1376616.1376746}, which contain billions of facts, is that they are not complete. This is due to the fact that these KGs represent diverse and complex real-world knowledge, which is often constantly changing and sometimes ambiguous \cite{mai2019-contextual}. The same applies to existing knowledge graphs in the cultural heritage domain. The extensively studied Knowledge Graph Completion (KGC)  methods are mostly embedding-based methods and are proven to achieve state-of-the-art link prediction performance \cite{nguyen2017survey, shen2022-comprehensive}. However, they follow the Closed World Assumption (CWA), which limits the entities to predefined sets. For cultural heritage KGs, where structured historical and cultural data are inherently incomplete and often hidden behind unstructured resources with high diversity, CWA will not be sufficient. Introducing new information to such KGs requires significant human expertise and effort. Thus, we argue that novel KGC methods under the Open World Assumption (OWA), where the prediction is not limited to known entities, are important to introduce new knowledge in KGs. We believe that methods involving multi-modal sources can help the extension of KGs under OWA in cultural heritage studies. 

To efficiently integrate these rich data sources to extend our KG, we propose a multi-modal KGC framework under OWA that uses the latest advancements of Large Language Models (LLMs) (eg, Llama 3 \cite{grattafiori2024llama3herdmodels}, GPT \cite{brown2020language}) and Vision-Language Models (VLMs) (eg, BLIP \cite{li2022blip}). Our methodology applies LLMs for semantic annotation and contextual understanding of textual data. At the same time, VLMs are used to analyze and interpret images. Although the outputs of both models are uncontrolled and may correspond to preexisting KG entities, we propose a grounding pipeline to ensure the predicted information is indeed both novel and accurate. We evaluate our proposed method by querying and testing the degree of reliability of both models, along with human evaluation. Our experiments show that this framework can efficiently extract knowledge from different data sources in a zero-shot manner, enabling the extension of KGs under the OWA. 

Finally, we show that our proposed pipeline can increase the number of entities in the KG by 20\%.  Our approach is domain-independent, as it could be applied to any KG and unstructured external data sources, including text and images, for KGC under OWA. In this context, with the purpose of facilitating KGC research, we benchmark our dataset on various existing KGC models. The results are in Appendix \ref{Apenx A}.

Our contributions are summarized as follows:
\begin{enumerate}
    \item We propose a novel cultural heritage KG: WJoconde, focusing on the French cultural heritage, with multi-modal data linked to heritage entities. WJoconde has more than 483k triples that aim to facilitate applications and research in the cultural heritage domain, especially for multi-modal research on KGs.
    \item We propose a novel KGC pipeline under OWA, which incorporates external information such as texts and images, along with an adapted evaluation method. By leveraging state-of-the-art LLMs and VLMs, our pipeline is able to extend a KG with new entities under OWA with high reliability. 
    \item We also aim to establish WJoconde a new KGC benchmarking dataset in the cultural heritage domain. We extensively evaluate various KGC models on WJoconde and its variants; the reported results can be found in Appendix \ref{Apenx A}. These results aim to encourage more research on domain-specific multi-modal KGC tasks. 
\end{enumerate}
To the best of our knowledge, this work is the first to create a comprehensive multi-modal benchmarking dataset in the French cultural heritage domain, combining textual and visual information. We further extend it with a proposed method under the open-world assumption along with an evaluation method to validate the results in such settings.

\section{Related work}\label{sec2}

\textbf{Unimodal knowledge graph completion methods.} Knowledge graph completion involves the prediction of missing entities or relations within a knowledge graph with or without external information. The most well-studied KGC models are embedding models. These models map entities and relations into an embedding space and then predict new triples using a scoring function applied to their embeddings. Such models involve translation models: TransE~\cite{DBLP:conf/nips/BordesUGWY13}, TransH~\cite{DBLP:conf/aaai/WangZFC14}, TransR~\cite{DBLP:conf/aaai/LinLSLZ15}, TransD~\cite{ji2015knowledge}, TorusE~\cite{ebisu2018toruse}, and Bilinear models: RESCAL~\cite{nickel2011three}, DistMULT~\cite{yang2014embedding}, HolE~\cite{nickel2016holographic}, ComplEx~\cite{DBLP:conf/icml/TrouillonWRGB16}, ComplEx\_n3~\cite{lacroix2018canonical}, TuckER~\cite{balavzevic2019tucker}.  
Embedding-based models are constrained by the fact that the scoring function relies on the embeddings of existing entities, preventing the addition of new entities to the KG.

\noindent\textbf{Multi-modal knowledge graph completion methods. } 
The majority of these models make use of structural data within the graph. However, existing knowledge graphs contain not only structural data but also textual and visual information. This has led to a surge of recent research interest in areas such as natural language processing for multimodal knowledge graphs (MKGs) \cite{chen2022hybrid}. Due to the inherent incompleteness of MKGs, approaches for their completion have been proposed \cite{chen2022hybrid,Shang_Zhao_Liu_Wang_2024}. These approaches involve the embedding of entities and relations according to the visual information attached to them. 
Some other works use large language models for knowledge graph completion, leveraging textual data as descriptions and name strings of entities and relations. \cite{yao2024exploring} is a preliminary work that explores different strategies for using LLMs for KG completion. This work demonstrates strong performance on textual data for this task. Combining this approach with our multimodal data is expected to improve results in this field. Moreover, there is growing interest in combining LLM with KGs, for example, \cite{peng2024refining} is the first to explore the use of LLMs for taxonomy cleaning, specifically for the noisy Wikidata taxonomy. Expanding on this, \cite{10.1007/978-3-031-33455-9_14} is an in-depth analysis of KG completion with LLMs. This analysis does not exploit the combined use of text and images. It demonstrates that the performance of the proposed methods depends on the specific properties targeted, an aspect of language model behavior that we have considered in this study.

\noindent\textbf{Cultural heritage knowledge graphs and linked open data. }
Many cultural institutions have adopted semantic web technologies to structure their data \cite{DBLP:conf/jcdl/TanTBODS21,carreiero2019-arco}. In 2012, the Amsterdam Museum published a book on the use of linked data. They start with XML data and present the process of converting it into linked form with a "man in the middle" perspective \cite{amsterdam-museum}. The Rijksmuseum, one of the first major museums to publish its collections according to Linked Open Data (LOD) principles (\cite{rijksarticle}) released a dataset of over 350,000 objects in March 2016. More broadly, LOD has become an widely adopted framework across cultural heritage collections, for example, Yale University, which developed a standards-based linked data discovery platform that unifies collections from its museums, libraries, archives, and special collections \cite{yaleLux2023}.
In parallel, several large-scale LOD initiatives have demonstrated how linked data can increase semantic expressivity and cross-collection research potential. These include the Zeri Photo Archive, converted to LOD through dedicated ontologies and CIDOC-CRM mappings \cite{10.1145/3051487}, the PHAROS consortium’s shared LOD platform for more than 20 million photographic records \cite{Caraffa_Pugh_Stuber_Ruby_2020}, the RKD Knowledge Graph integrating extensive Dutch art-historical documentation \cite{rkdkg}, and the Golden Agents project, which connects heterogeneous biographical and artistic datasets from the Dutch Golden Age \cite{DBLP:conf/bd/BrouwerN17}. Together, these efforts highlight the maturity of LOD infrastructures in cultural heritage and motivate the need for complementary approaches such as multimodal KG extension under the open-world assumption.

In their survey \cite{FIORUCCI2020102}, Fiorucci et al. list publicly available datasets used for machine learning experiments in cultural heritage, but only one (Europeana) includes relationships. In \cite{10.1007/978-3-030-11012-3_53}, the authors aim to assign a restricted number of tags to parts artwork images, using a large image model trained with natural photographs. They do not use other data sources and, contrary to our objective, have a limited number of possible values for the targeted properties.

\noindent\textbf{Knowledge graph datasets for KGC. }
Common databases that are frequently used in knowledge graph literature include FreeBase, a versatile structured data repository, and WordNet \cite{10.1145/219717.219748}, an extensive English lexical database known for its rich word relationships and semantic meanings. Several KG datasets are built from these two bases, such as FB15k \cite{DBLP:conf/nips/BordesUGWY13}, FB15k237 \cite{toutanova-etal-2015-representing}, WN18 \cite{DBLP:conf/nips/BordesUGWY13}, and WN18RR \cite{dettmers2018convolutional}. These datasets are commonly used to evaluate various KGC tasks, mainly with unimodal approaches. FB15k-237-IMG (14,541 entities and 237 relations) \cite{chen2022hybrid} and WN18-IMG (40,943 entities and 18 relations) \cite{DBLP:conf/nips/BordesUGWY13} are extensions of these datasets with images, where each entity is linked to 10 images. These multimodal datasets are used to evaluate recent embedding-based approaches for multimodal KGC.

\section{The WJoconde dataset}\label{sec3}

In this section, we describe the creation of WJoconde, a novel knowledge graph for the French cultural heritage, built from the subset of Wikidata entities linked to Joconde database \cite{moissinac_jean_claude_2020_3986498}. 

The creation of WJoconde followed a multi-step semi-automated pipeline. First, we extracted a context graph from Wikidata by selecting all entities carrying the Joconde ID (P347). This step leverages the fact that Wikidata already contains curated mappings from Joconde records to structured semantic information. Second, we cleaned and normalized this extracted graph by removing redundancies and low-quality relations. Third, we enriched each entity with external text and images through automated retrieval from open heritage sources, Wikimedia Commons, and Wikipedia. All steps were implemented using Python scripts interacting with the Wikidata API and WDumper exports.

We use WJoconde as the raw dataset, and further constructed two variants: WJoconde\_cleaned and WJocondeMM. WJoconde\_cleaned is a filtered version of the raw dataset, following the procedure used in FB15k-237 \cite{toutanova-etal-2015-representing} and WN18RR \cite{shang2018endtoend}, in order to remove inverse, redundant, and extremely sparse relations that can artificially inflate KGC performance and hinder meaningful model evaluation. WJocondeMM is another filtered version of the raw dataset, where we select the entities that both have text descriptions and visual images, making it a smaller, full multi-modal dataset. Below, we present an overview of the original Joconde database, then describe the steps for building WJoconde and its two variants, while justifying the design choices.


\subsection{Joconde database}
The Joconde database is operated by the French Ministry of Culture. At the time of creating this work, it provides metadata on some 600,000 creations from French cultural heritage. It is mainly fed by data from the curators of the museums where these creations are housed. As such, it constitutes a high-quality reference dataset for the field of French cultural heritage. An extract from the Joconde database is available in the form of a JSON file\footnote{\url{https://www.data.gouv.fr/fr/datasets/collections-des-musees-de-france-base-joconde/} (checked on 17/02/2025)}.

The Joconde database is used by Wikidata contributors to inject the description of certain works into Wikidata. This is done either by manual intervention or by "robots", as defined in the Wikidata platform environment. The data model is that of the works represented in Wikidata. As of $19/02/2025$, there are  $24,275$ works of Joconde in Wikidata.

The Joconde database was used in one of our projects to create a knowledge graph entitled SemJoconde. The data model used was based on CIDOC-CRM. When creating SemJoconde, considerable effort was devoted to linking the concepts from the Joconde database to those used in Wikidata \cite{moissinac2020semjoconde}. SemJoconde can be queried on a SPARQL access point\footnote{\url{http://datamusee.r2.enst.fr/\#/dataset/SemJoconde/query} (checked on 18/02/2025)}. This work enabled us to create a robot to inject the works of Joconde into Wikidata, taking advantage of the quality of the links produced for SemJoconde. 

The conditions for injecting Joconde into Wikidata ensure that the quality of the data in the Joconde database is preserved in the corresponding Wikidata entities.

During this project, the Joconde database did not provide any images related to the works of art described. As Wikidata presented works from the Joconde database linked to other databases, we turned to Wikidata's data. Moreover, we did not use SemJoconde as a source dataset because it does not contain links to images and the CIDOC-CRM-based graph structure complicates its use; a work is represented by three different entities: the act of creation, the physical object, and the conceptual object. A more direct representation is used in the Wikidata model. In addition, Wikidata data benefits from a user validation system.

\subsection{WJoconde: raw dataset}
In \cite{mimouni2019knowledge,mimouni2024-context}, the authors introduced the concept of a context graph, a graph representing knowledge on a specific topic, typically extracted from a larger, more general graph. The purpose is to work with a smaller, more focused graph, containing a subset of relations and possible values for each property. Various works have since used similar concepts, such as \cite{DBLP:journals/corr/abs-1906-04536,nguyen2022wikidatalite}.

In the work presented here, we have established a context graph using the WDumper tool\footnote{\url{https://wdumps.toolforge.org/} (checked on 17/02/2025)}. Works in the Joconde database present in Wikidata generally fill in the property \texttt{Joconde work ID} ($P347$) with the identifier of the work in the Joconde database.  
We've configured WDumper to retrieve all triples directly linked to entities, identified by their Wikidata IDs (QID), with a $P347$ property, i.e., all data concerning entities with a Joconde ID (see \footnote{\url{https://datamusee.wp.imt.fr/2023/05/31/extrait-de-wikidata-oeuvres-de-la-base-joconde/} (checked on 17/02/2025)} for further details). In the following and in related documents, we use  WikidataJoconde or its short name WJoconde interchangeably to refer to the same dataset.
While WJoconde covers only a subset of Joconde, this is a deliberate choice to focus on entities that have passed several validation steps, namely creation by museum curators, integration into Wikidata, and controlled property mapping.

\subsubsection{Main types} 

As we are working with data extracted from Wikidata, the data model employed is a subpart of the model adopted by Wikidata. The entities presented are typed using the $P31$ (\texttt{instance} of) property. There are $303$ types in WJoconde.  Table \ref{tab1} presents the main types of heritage objects in WJoconde. All types beyond \texttt{statue} have $161$ instances or fewer, and there are $14491$ instances of painting. Note that some entities can be both subjects and objects, for example when two works are linked as companion pieces in a diptych.

\begin{center}
\begin{table}[h]
\caption{Main types (property $P31$ - Instance of)} 
\label{tab1}
\begin{tabular}{|l|l|l|l|l|}
\hline
URI &  Label & Entities count & Average number of properties & Properties count \\
\hline
wd:Q3305213 & painting & 14491 & 26.25 & 380360 \\
wd:Q860861 & sculpture & 1884 & 18.07 & 34050 \\
wd:Q220659 & archaeological artifact & 1010 & 24.85 & 25014 \\
wd:Q179700 & statue  & 225 & 26.02 & 5855 \\
\hline
\end{tabular}
\end{table}
\end{center}

\subsubsection{Main properties} 
The paintings described in WJoconde can refer to $215$ different properties. Table \ref{tab2} shows the most frequent ones and they are similar across different heritage types. In WJoconde, there are 370 properties in total, which are also relations or edges in the KG.

\begin{center}
\begin{table}[h]
\caption{Main properties of paintings}\label{tab2}
\begin{tabular}{|l|l|l|}
\hline
URI &  Label & Instance\\
     &       & count\\
\hline
wdt:P180 & depicts & 58261 \\
wdt:P186 & made from material & 28458 \\
schema:description & description & 27622 \\
rdfs:label & label & 21317 \\
wdt:P217 & inventory number & 19006 \\
wdt:P195 & collection & 16119 \\
wdt:P31 & instance of & 14750 \\
wdt:P170 & creator & 14600 \\
rdfs:type & type & 14597 \\
wdt:P347 & Joconde work ID & 14550 \\
wdt:P276 & location & 14473 \\
wdt:P2048 & height & 13623 \\
wdt:P2049 & width & 13597 \\
wdt:P571 & inception & 13571 \\
wdt:P6216 & copyright status & 11954 \\
wdt:P136 & genre & 11008 \\
wdt:P1476 & title & 9502 \\
wdt:P18 & image & 9026 \\
wdt:P9394 & Louvre Museum ARK ID & 5771 \\
wdt:P973 & described at URL & 5128 \\
wdt:P1257 & depicts Iconclass notation & 4964 \\
wdt:P528 & catalog code & 3498 \\
wdt:P1212 & Atlas ID & 3248 \\
wdt:P127 & owned by & 3078 \\
skos:altLabel & alternative label & 3022 \\
\hline
\end{tabular}
\end{table}
\end{center}

\subsubsection{Statistics}

Table \ref{tabstat} presents the basic statistics of WJoconde. Table ~\ref{t1} also shows the characteristics of relations in WJoconde where $1-1$, $1-n$ denote one-to-one and one-to-many relations, respectively, and $n-1$ and $n-n$ are defined similarly. One-to-one relations connect one entity to a unique entity, one-to-many relations connect one entity to several others, many-to-one relations link multiple entities to a single one, and many-to-many relations represent multi-directional connections.

To determine the cardinality type of each relation $r$, we compute two standard metrics: the average number of tail entities per head entity (tph) and the average number of head entities per tail entity (hpt). Specifically, $\text{tph}(r)$ is calculated as the mean number of distinct tail entities associated with each unique head entity under relation $r$, while $\text{hpt}(r)$ is defined analogously as the mean number of distinct head entities per tail entity. Following prior work, we apply a threshold of 1.5 to classify relations into cardinality categories: a relation is considered one-to-one if $\text{tph} < 1.5$ and $\text{hpt} < 1.5$, one-to-many if $\text{tph} \geq 1.5$ and $\text{hpt} < 1.5$, many-to-one if $\text{tph} < 1.5$ and $\text{hpt} \geq 1.5$, and many-to-many if $\text{tph} \geq 1.5$ and $\text{hpt} \geq 1.5$.

An example of the WJoconde can be found in Appendix B.

\begin{center}
\begin{table}[h]
\caption{WJconde dataset characteristics } \label{tabstat}
\begin{tabular}{|l|l|l|}
\hline
Element & Count\\
\hline
triples & 483,938  \\
entities as subject & 28524 \\
entities as object & 37708  \\
distinct entities & 61383 \\
entities in subject pos. and object pos. & 4,491 \\
entities sharing at least one depicts & 13,916 \\
entities sharing at least one genre & 11,170 \\
relationships & 370  \\
\hline
\end{tabular}
\end{table}
\end{center}

\subsection{WJoconde\_cleaned: the cleaned dataset} \label{cleandata}

In \cite{Galarraga2013Amie}, for each relation in a graph the authors define a measure relation called "functionality", and its inverse, "inverse functionality". When either measure reaches its maximum value of 1.0, both ends of the relation can be used interchangeably, effectively identifying the same entity. In \cite{toutanova-etal-2015-representing} and \cite{shang2018endtoend}, the authors follow a similar path to remove redundant relations and rare entities. Below, we show  how these principles can be used to obtain a filtered version of our original dataset.

\begin{enumerate}
    \item Remove near-duplicate and inverse relations. Inspired by \cite{toutanova2015observed}, a near-duplicate relation is defined as whether the set of entity pairs in the relations is either almost the same (e.g. $97\%$ of the pairs are in the intersection), an inverse relation is defined as whether the set of inverse entity pairs involved in a relation is almost the same. Near-duplicate and inverse relations essentially convey the same information, and thus, we retain only one of a pair of inverse or duplicate relations.
    \item Remove ID-related relations. We automatically select all relations that have "ID" in their names, then remove all triples involving these relations.
    \item Remove URL-related relations. We automatically select all relations that have "URL" in their names, then remove all triples involving these relations.
    \item Manual selection of relations. We manually check the remaining relations that are ID and URL-related, but are not found in the previous step, such as relation P8091 (Archival Resource Key), which is an ID number, and $P4896$ (3D model), which is a URL to a 3D model. We also remove the relations with low frequency in the dataset, as some involve only $1$ triple in the entire dataset and thus contribute little information to the graph.
\end{enumerate}

Following these procedures, we derived a more compact dataset featuring the most pertinent relations, and we named this refined dataset WJoconde\_cleaned. Details about the dataset can be found in Table \ref{t1}.

\subsection{WJocondeMM: the multi-modal dataset}
As mentioned earlier, WJoconde is a French cultural heritage dataset where most of the heritage objects have detailed text descriptions and clear visual images. They are valuable components as they offer a multidimensional perspective while providing more historical, cultural, and contextual insights, which is useful both for applications such as museum query systems or future research. 

For each entity in WJoconde, we aimed to collect more contextual information. To achieve this, we created an enriched multi-modal dataset by applying the following pipeline. All these steps were performed automatically using public APIs, with the entity identifiers serving as stable keys to retrieve the corresponding text and image resources:
\begin{enumerate}
    \item We collect attributes from the Open Heritage Platform\footnote{\url{https://www.pop.culture.gouv.fr/} (checked on 17/02/2025)} in France using the Joconde ID. These attributes are transformed into text descriptions following templates that include the heritage name and the source text, with URLs removed. Furthermore, most entities in WJoconde have a Wikipedia page; we collect all text descriptions from each entity’s Wikipedia page as an additional source of contextual information. Finally, we created $2$ French and English versions of contextual data.
    \item We collected images related to each entity from Wikimedia Commons\footnote{\url{https://commons.wikimedia.org/wiki/Main_Page} (checked on 17/02/2025)} for each entity in WJoconde. This matching process was fully automated because both Wikidata and Wikipedia rely on the same stable entity identifiers, ensuring that the retrieved images are highly accurate and relevant without requiring additional quality verification.
    \item We selected the entities that have both a text description and an image.
\end{enumerate}

\begin{figure}[htp]
    \centering
    \includegraphics[width=0.9\linewidth]{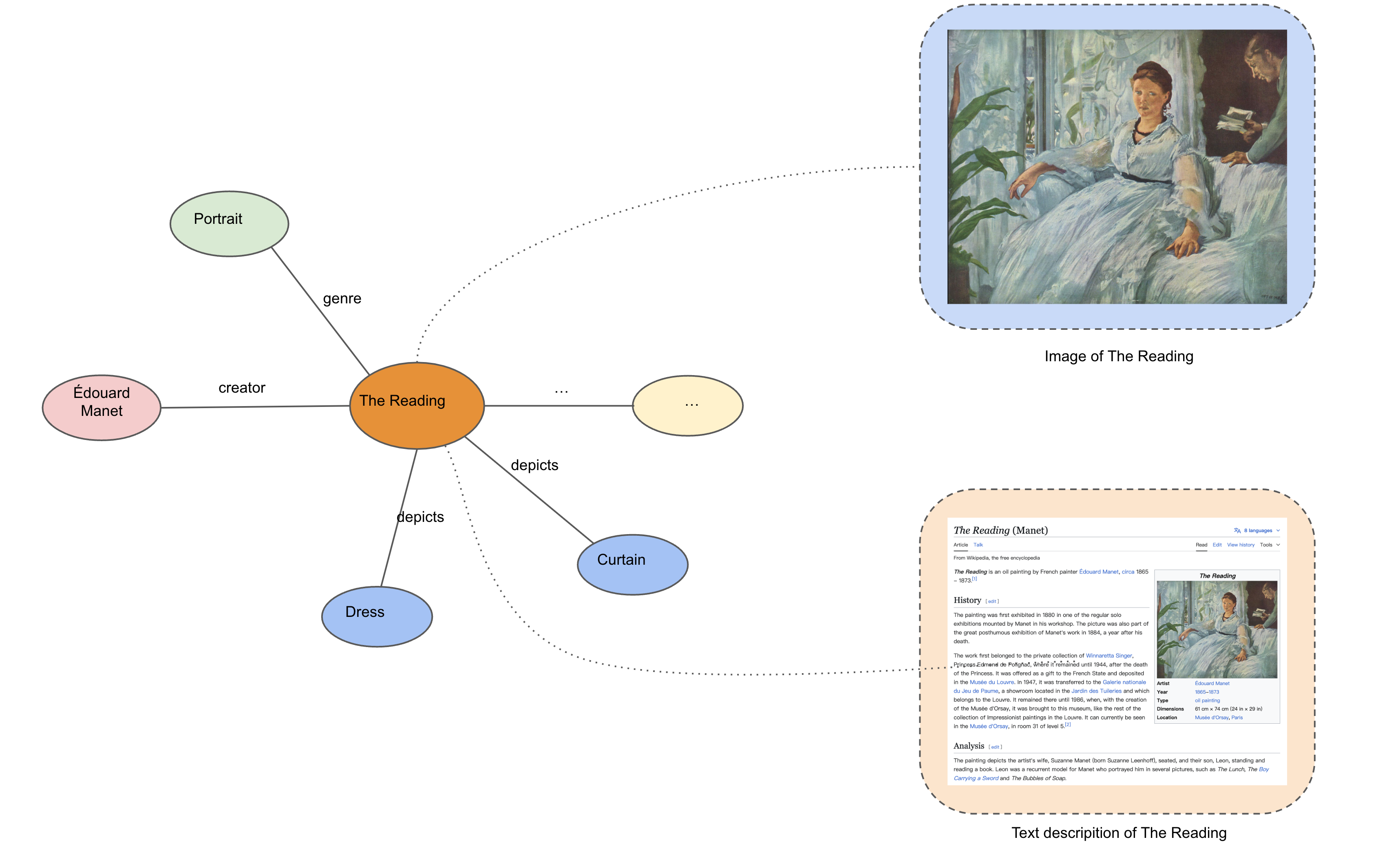}
    \caption{Example of the WJocondeMM}
    \label{tr}
\end{figure}

We call this dataset WJocondeMM, and we use WJocondeMM\_en for the English version and WJocondeMM\_fr for the French version, where each entity in the KG can be connected to a text description and a visual image, as shown in Fig. \ref{tr}. Details about the dataset can be found in Table \ref{t1}.

\begin{table}[!htp]
    \centering
    \caption{Summary of the datasets}
     \label{t1}
    \begin{tabular}{llllllll}
    \hline
        Dataset & Triples & Entities & Relations & 1-1 & 1-n & n-1 & n-n \\ \hline
        WJoconde & 483,938 & 175,822 & 370 & 294 & 14 & 55 & 7 \\ 
        WJoconde\_cleaned & 93,218 & 372,494 & 121 & 64 & 7 & 46 & 4 \\ 
        WJocondeMM\_en & 19,122 & 7,749 & 121 & 72 & 11 & 32 & 6 \\ 
        WJocondeMM\_fr & 240,935 & 60469 & 121 & 66 & 9 & 42 & 4 \\
            \hline
    \end{tabular}
\end{table}

\subsection{Summary and Quality of the Dataset}
The Joconde data integrated into Wikidata is of very high quality. As previously noted, the work of Moissinac et al. \cite{moissinac2020semjoconde} established verified semantic alignments between the Joconde vocabulary and Wikidata, ensuring reliable cross-referencing between the two resources. In addition to this curated alignment effort, Wikidata benefits from two major quality-enhancing mechanisms:
\begin{itemize}
\item a large community of human contributors who continuously verify, correct, and enrich the data,
\item automated bots that regularly inspect Wikidata to enforce consistency constraints and complete certain properties.
\end{itemize}

Taken together, these mechanisms ensure that the portion of Joconde represented in Wikidata, on which WJoconde is built, is maintained with strong reliability and semantic coherence.

Overall, the WJoconde dataset provides a high-quality, domain specific cultural heritage knowledge graph derived from curated heritage information and enriched with multimodal data. Through the cleaning steps, we obtained a compact yet semantically expressive graph, suitable for robust evaluation of knowledge graph completion models. Through multimodal enrichment, each entity is accompanied by at least one textual and one visual resource, enabling research in both unimodal and multimodal KGC and offering a rich foundation for downstream cultural-heritage applications.

\subsection{A multi-modal benchmark for KGC}
We also want to make this dataset a new benchmark for KGC models. Existing benchmarks such as FB15k and WN18 capture only simplified subsets of knowledge bases and therefore lack the operational complexity, ambiguity, and incompleteness characteristic of large curated collections. This highlights the need for benchmark datasets that better reflect practical application scenarios across domains and include multimodal information to support more advanced models. We apply several common KGC models to our datasets and report their performance in Appendix A. Our selection covers Translation models, Bilinear models, and Neural Network models, and we report Mean Rank (MR), Mean Reciprocal Rank (MRR), and Hit@10 scores across three variants of the WJoconde dataset. We find that translation-based models have difficulty accurately modeling the KG. This is possibly due to the complexity of our dataset, where the relation between entities is more complex (more 1-to-N, N-to-1 or N-to-N relations), especially for the TransR model. This result is consistent with many works that point out the limitations of such models. Bilinear models such as ComplEx and DistMULT work relatively better due to their increased expressiveness through matrix factorization. The best-performing model is TuckER, a deep neural model capable of representing complex relations. Our findings indicate that completing knowledge graphs with domain-specific applications requires more complex models capable of capturing additional information from the KG. Performance is expected to improve further when combined with multi-modal methods.

\section{Open World Knowledge Graph Extension}\label{sec4}

In this section, we discuss our proposed knowledge graph extension pipeline that can introduce new information to the KG, especially in areas such as cultural heritage, where rich text and visual images are available.

Knowledge graph completion is one of the major topics related to KG research. So far, most of the KGC models have been embedding-based, for example, TransE \cite{DBLP:conf/nips/BordesUGWY13} and RESCAL \cite{nickel2011three}. These involve the mapping of entities and relations to an embedding space, followed by the prediction of new triples using a scoring function applied to the embeddings of these new triples. As a result, only preexisting entities and relations in the KG can be predicted, which follows the CWA. 

However, in cultural heritage KGs, predicting new entities is crucial, as existing KGs often lack sufficient entities due to the evolving nature of historical research and the diversity of heritage information. Much of the detailed information about an object, for example, its features, historical significance, and contextual relationships, is contained in text descriptions and visual images. These sources frequently contain unique information specific to particular objects, which might not be widely recognized or previously documented in structured databases. 

\begin{figure}[htp]
    \centering
    \includegraphics[width=1\linewidth]{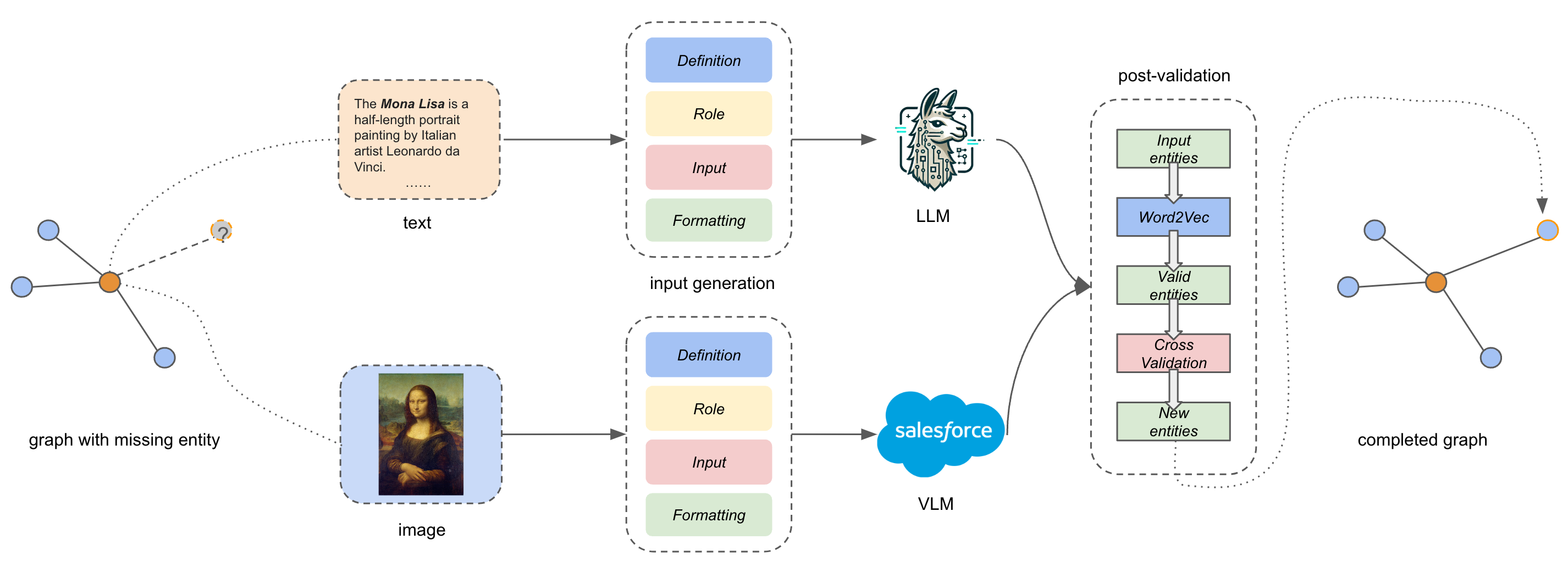}
    \caption{Pipeline of our approaches.}
    \label{app2}
\end{figure}

To exploit this information for extending WJoconde under OWA, we propose a new KGC method that takes advantage of the recent LLMs, eg, GPT-3 \cite{brown2020language} and Llama 3 \cite{grattafiori2024llama3herdmodels}, and visual-language models such as BLIP \cite{li2022blip}. By leveraging unstructured information such as text and images related to each entity, the models predict new information that is not necessarily already in the KG. Fig. \ref{app2} shows the pipeline of our method. 

We define our method in three stages:

\textbf{Information collection:} Given a triple with a missing entity, we first collect related text descriptions and image visualization of any related entities or relations. For WJoconde, we collect rich text and image information as described in Section \ref{sec3}.

\textbf{Entity extraction:} Then, we apply LLM and VLM to process the given information respectively. To do this, we design prompt templates that involve the known entity and relation in the triple with their unstructured information and require the model to output JSON formats of answers. We demonstrate that this approach is both efficient and effective for the most popular open-source LLMs and VLMs, without requiring any model training. We attribute this to the fact that these LLMs are trained, among other data sources, on Wikidata. At the end of this stage, each model will give a list of predicted entities that serves as a candidate for our query. We also observe that, due to the strong ability of the latest models, they can output answers not seen in the given text or image from their inner knowledge.

\textbf{Joint validation:} One problem related to LLMs and VLMs is their outputs are not controlled by any list of "true values", and for Open World Knowledge Graph Completion, the challenge is how to be sure the newly introduced entity is indeed new, rather than synonyms of an existing entity. To resolve this challenge, we jointly validate the outputs of both LLM and VLM. We conduct an \textbf{Entity Matching} validation step by leveraging a simple Word2Vec \cite{mikolov2013efficient} model to calculate the similarity between LLM-generated answers and VLM-generated answers, and also between both model-generated answers and existing entities in the KG, and then select the answers that are above a certain threshold. At the end of this stage, we obtain new information that is unique for both models and new for our KG.

Another challenge for open-world KGC is how we evaluate the newly introduced information. The traditional ranking-based methods don't fit in this case since they require knowing the true value in the KG. Intuitively, after the entity-matching validation step, we can assign high confidence to the outputs of the LLM and VLM, as these models have demonstrated strong performance in information extraction tasks following large-scale pre-training and task-specific fine-tuning. However, the reliability of the answers generated by both models remains an open question. To achieve this, we designed an extra \textbf{Model Validation} stage, where we use the same prompt template and the same external texts and images, but rather than ask the model to predict new information, we ask the model whether the existing entities in our KG are indeed related to the tested object. By doing so, we assess the reliability of the LLM and VLM in entity recognition from the provided information. This is further supported by our subsequent extensive human evaluation.

\section{Experiments}\label{sec5}

In our experiments, we try to complete WJocondeMM\_en for demonstration as it includes both text and image information for properties such as "time", "creator", "depicts" and "genre". One of the main properties we aim to complete in WJoconde is "depicts", which describes the content or features of a heritage object. The entities associated with this property are far from complete, as a single object may relate to an arbitrary number of depicted elements. For the LLM and VLM in our method, we use the pre-trained LLaMA 2 7B and BLIP Large models, respectively. However, this choice is not restrictive, and other LLMs and VLMs can be integrated into the pipeline. Our experiments show that using the pre-trained LLM and VLM can accurately extract new entities in a zero-shot manner. We also fine-tuned these models, but no major improvement was shown.

\subsection{LLM and VLM for entity extraction}
LLMs and VLMs show strong zero-shot capability in various tasks; however, their predictive nature means the outputs are unstructured and occasionally prone to hallucinations. To help the models better understand the task and control the output format, we designed prompt templates for both models, following several design principles:

\begin{enumerate}
    \item Definition: give the definition of knowledge graphs and clearly describe the knowledge graph completion task, so the LLM and VLM can have a clear context to give a desired output.
    \item Role: give a role to the LLM and VLM in the KGC scenario, specify the way we want to process the input and the specific property we aim to complete, in order to enhance the robustness of the responses.
    \item Input: give the input data that we want the LLM or VLM to process, either text or an image, from which we aim to extract information. 
    \item Formatting: define a certain format of the output, in order to constrain the response from both models, such as file format, response length, etc. Here, we set the output to JSON format with short words as answers.
\end{enumerate}

\begin{figure}[htpb]
    \centering
    \includegraphics[width=.8\linewidth]{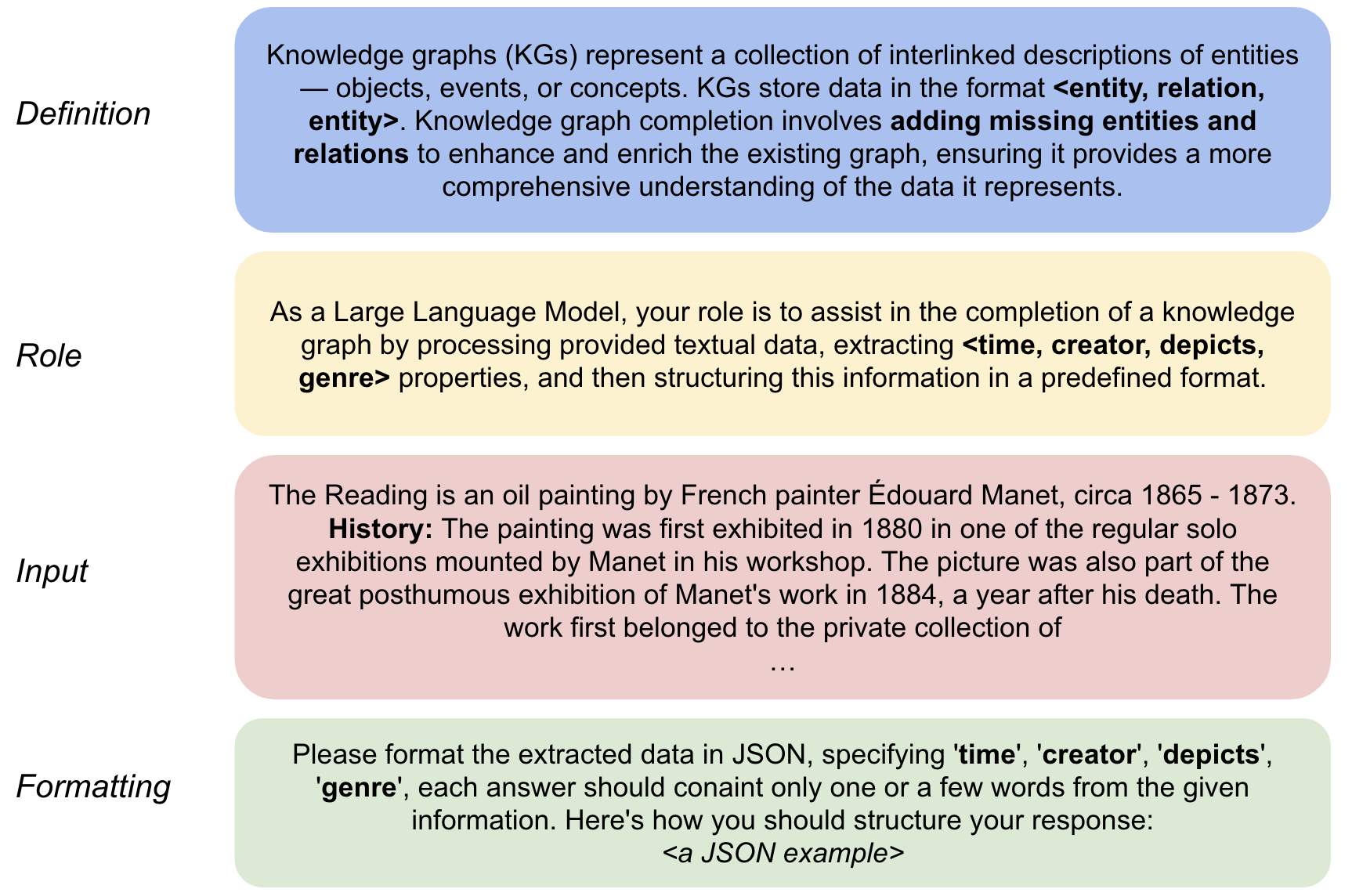}
    \caption{Prompt example for LLM}
    \label{llm}
\end{figure}

\begin{figure}[htpb]
    \centering
    \includegraphics[width=.3\linewidth]{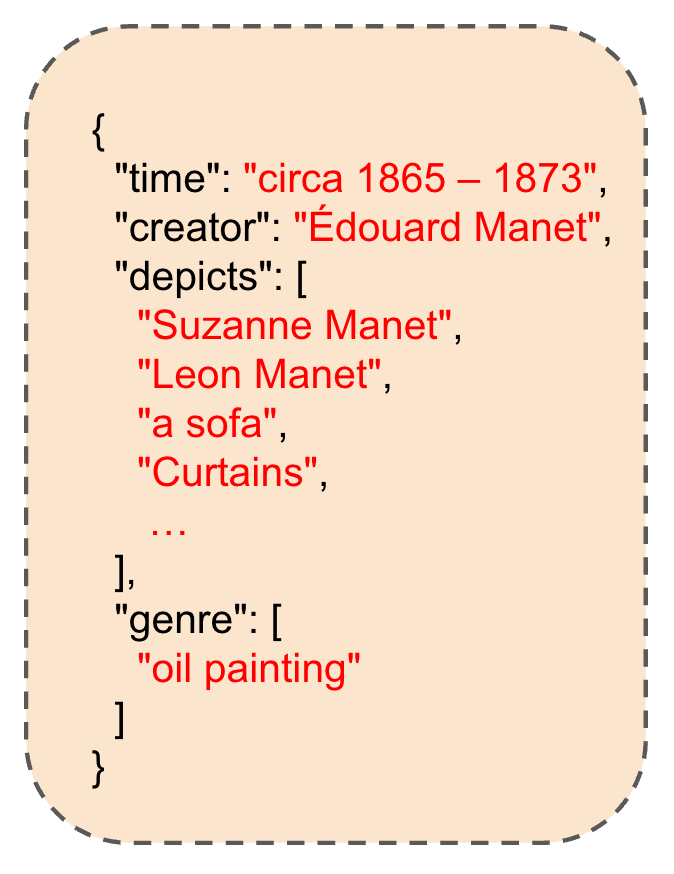}
    \caption{Output from LLM}
    \label{out}
\end{figure}

Following this template, we construct a prompt for each entity we aim to process by specifying the task and text input. An example is shown in Fig. \ref{llm}. The LLM outputs answers in JSON format, allowing the results to be easily processed. An example is shown in Fig. \ref{out}.

\begin{figure}[htpb]
    \centering
    \includegraphics[width=.8\linewidth]{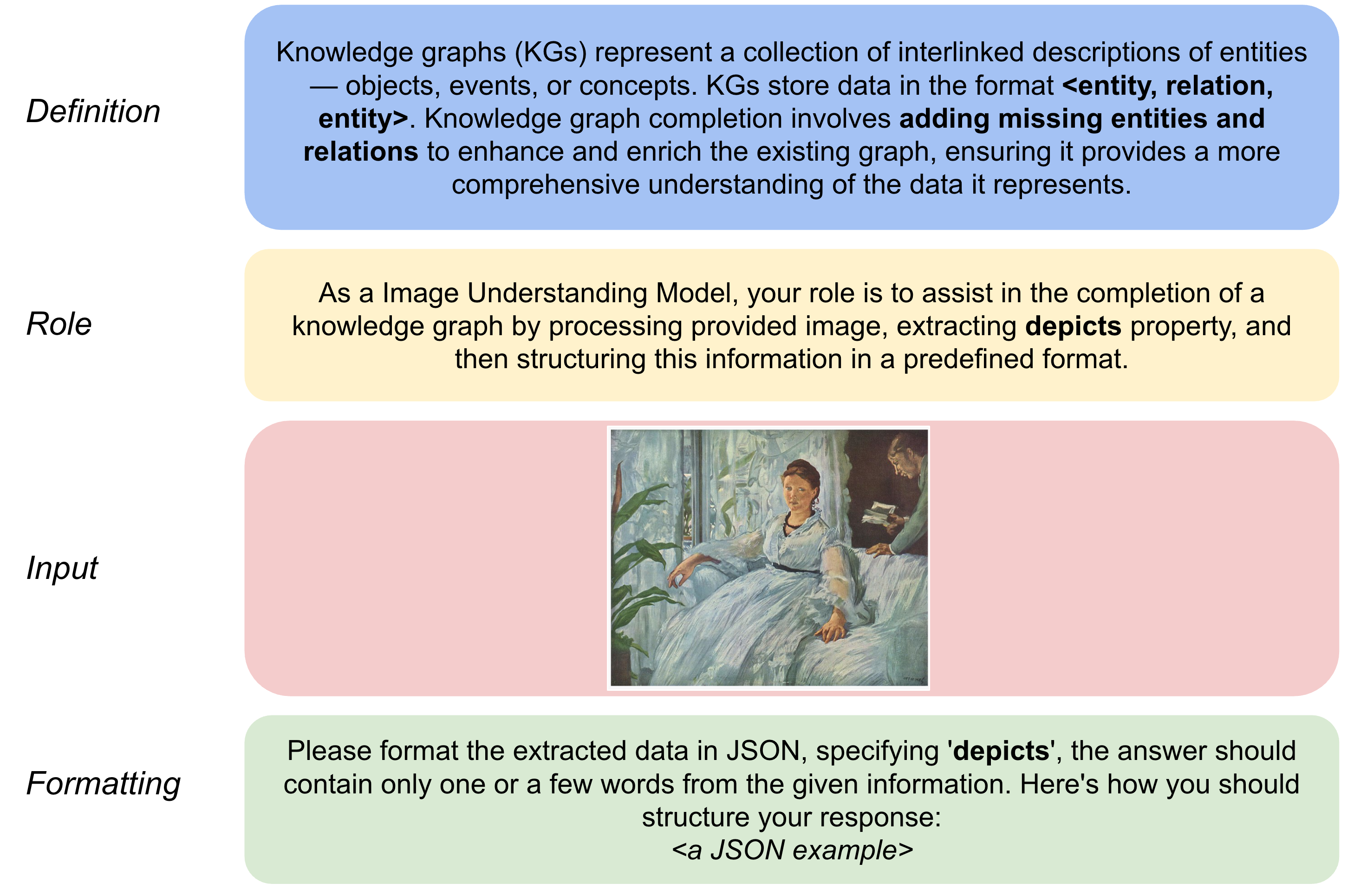}
    \caption{Prompt example for vision-language model}
    \label{ima}
\end{figure}

\begin{figure}[htp]
    \centering
    \includegraphics[width=.2\linewidth]{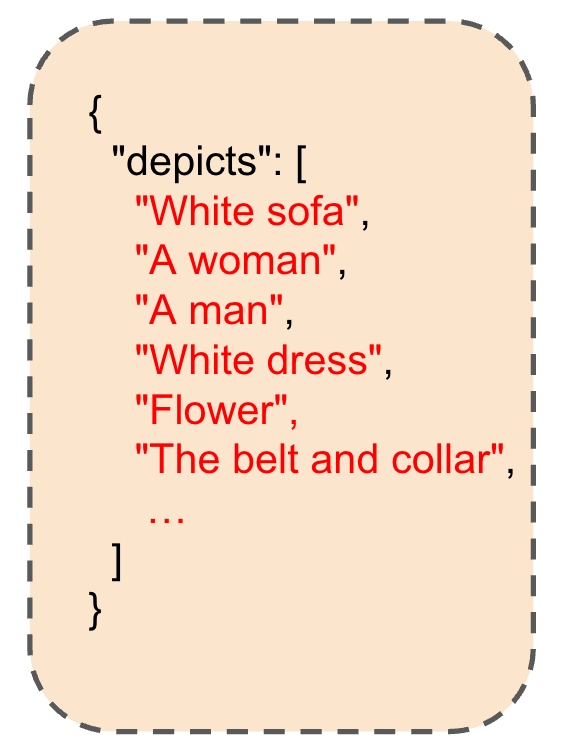}
    \caption{Output of vision-language model}
    \label{imgjson}
\end{figure}

Similar to the LLM, given the template and an image as input, as shown in Fig.\ref{ima}, the VLM will give the result in a JSON file. Fig. \ref{imgjson} shows the result of this example.

All prompts were executed in a controlled Python framework using HuggingFace Transformers for both LLM and VLM. Prompts were generated automatically by a templating module and delivered to the models through batch-processing scripts. Each output was parsed using a JSON schema validator that ensured conformity with the requested output format. Invalid generations were re-queried once with a stricter prompt.

\subsection{Validation and evaluation}
The raw outputs from the LLM and VLM could fall into 3 types: answers already in the KG; answers that are synonyms to the ones in KG or between the models, and answers that are new to the KG. Thus, we perform further post-validation to select the new answers to the KG.

\subsubsection{Entity matching}
In order to select new entities, we first filter out preexisting entities or synonyms. In our example, Curtain is already in our KG and Curtains is the output. In this case, we do not want to add Curtains to the KG.

\begin{figure}[htp]
    \centering
    \includegraphics[width=.8\linewidth]{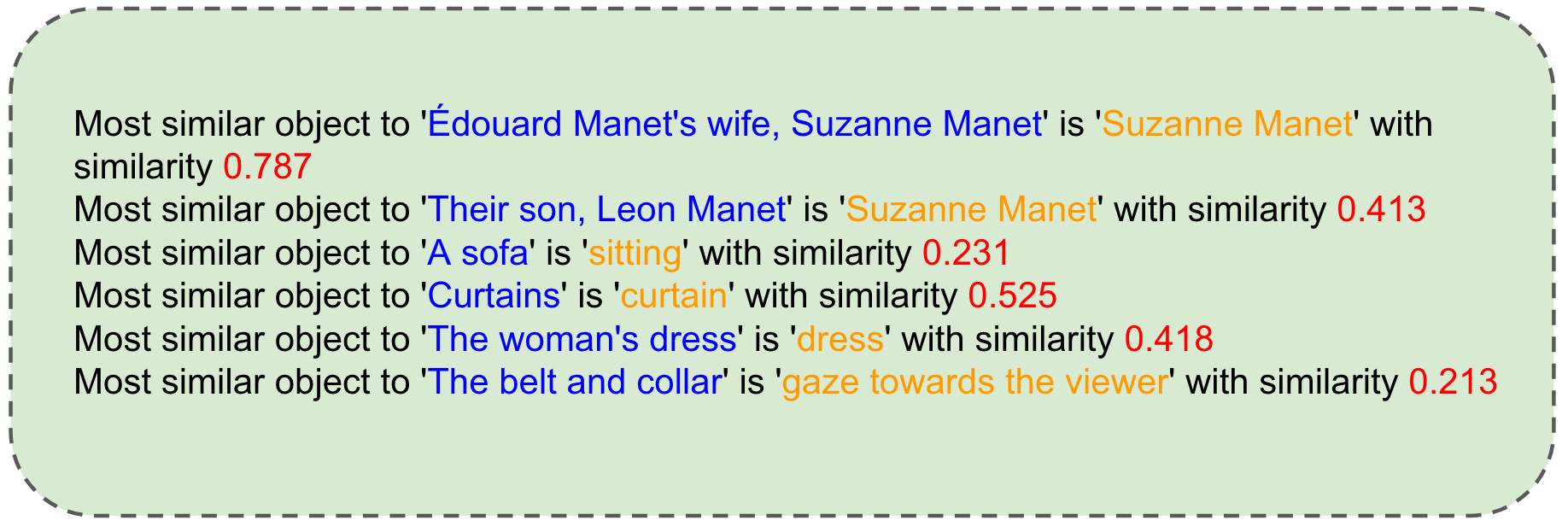}
    \caption{Similarity between entities.}
    \label{output3}
\end{figure}

To match the extracted entities and the entities in the KG, we apply a Word2Vec model to calculate word similarity between the LLM or VLM generated entities and the entities already in the KG.We set a threshold $\tau$ for the similarity score for a stricter filtering. For a given entity and a relation, we compute the similarity with all entities in the KG that relate to the relation. If the similarity is less than $\tau$, we consider it to be a new entity. We selected $\tau = 0.4$ by randomly testing 200 predictions, and we found 0.4 has the best performance to avoid finding synonyms. An example is shown in Fig. \ref{output3}.

We also validate the outputs between the two models, in order to avoid synonyms between the outputs of the two models. In the end, we have two sets of answers with entities that are new to the KG and unique for the combined set.

\subsubsection{Cross validation}

One additional step to further improve the quality of the generated answers is to have the LLM and VLM validate each other's outputs, especially for properties like "depicts", which can be easily validated with visual images. A common issue when using LLMs for extraction tasks is the inconsistency between answers and tasks, or even hallucinations. To address this, we can transform the task to a VLM QA task, which typically has higher accuracy, to further filter out low-confidence answers. Only entities verified as present by the image model are retained. This increases precision by removing hallucinations that are not visually supported.

\begin{figure}[htp]
    \centering
    \includegraphics[width=.3\linewidth]{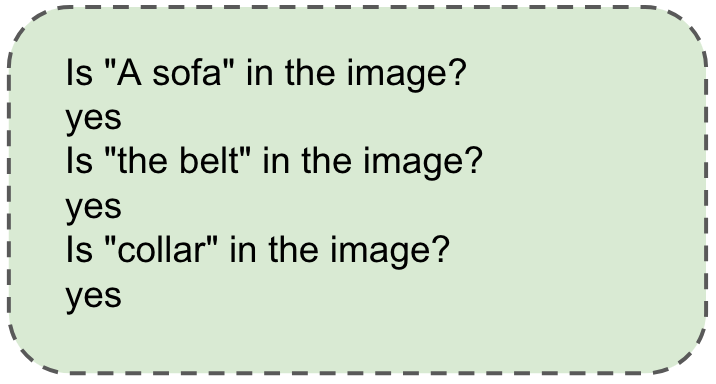}
    \caption{Validation from the vision-language model.}
    \label{ev}
\end{figure}

We perform this validation by asking the vision-language model whether the new entity given by the LLM is in the image. Fig. \ref{ev} provides an example. Following the entity matching, the VLM validated the presence of the new entity, thus, we can add "the belt" and "collar" and "the sofa" to our KG with a higher confidence.

\subsubsection{Evaluation} Traditional evaluation metrics for knowledge graph completion, such as Hit@10, Mean Reciprocal Rank (MRR), and Mean Rank (MR), focus on predicting links between existing entities and are not fully suitable under the OWA. 

To assess the quality of the entities generated by our method, we first evaluate the credibility of the LLM and VLM. To do this, we test the confidence of both models in accurately recognizing entities already in our KG, given the text or image information. Table \ref{eva} shows the results. For the LLM and VLM combined, they can recognize 90\% of the entities following our pipeline, showing high credibility for both models.

To further assess the quality of the newly extracted entities, we conducted a human evaluation. Two annotators independently reviewed all generated entities that relate to 606 artworks with "depicts" relation in our experiment and labeled each as either correct or incorrect. For each annotator, we computed the proportion of entities judged correct, and we report the average of the two annotators’ scores. Table \ref{eva2} presents the resulting human-judged precision.

\begin{table}[h]
    \centering
    \caption{Evaluation results on recovering ground truth}
    \begin{tabular}{llll}
    \hline
        models & Recall & F1 score \\ \hline
        Llama & 0.722 & 0.839 \\ 
        BLIP & 0.848 & 0.918 \\ 
        Llama + BLIP& 0.902 & 0.948 \\ 
        \hline
    \end{tabular}
    \label{eva}
\end{table}

\begin{table}[h]
    \centering
    \caption{Human evaluation on newly generated responses}
    \begin{tabular}{llll}
    \hline
        models &  Precision \\ \hline
        Llama &  0.864   \\ 
        BLIP & 0.905\\ 
        Llama + BLIP&  0.920\\ 
        \hline
    \end{tabular}
    \label{eva2}
\end{table}

\subsubsection{Results}
In this section, we present the results of applying the above pipeline to our WJoncodeMM\_en knowledge graph, as summarized in Table \ref{tab:soge}.  After applying the LLM and VLM for entity extraction, and performing an initial filtering of similar entities between the two models, they generated 6,431 entities, corresponding to roughly 89.80\% of the number of the entities in the raw dataset. 
After the post-validation steps, entities similar to those already in the dataset are filtered out, resulting in 4,394 new final entities to be added to the dataset, with 68.32\% filtered entities among all generated entities. In the end, our approach introduces 61.36\% more entities, which we consider to be a very encouraging result.

In Table \ref{tab:stat}, we present the statistics of the raw KG and the extended KG using our proposed approach. We extended the original WJocondeMM\_en by 56.7\% more entities, and triples in the KG are extended from 19,112 to 23,516.

\begin{figure}[h]
    \centering
    \includegraphics[width=.8\linewidth]{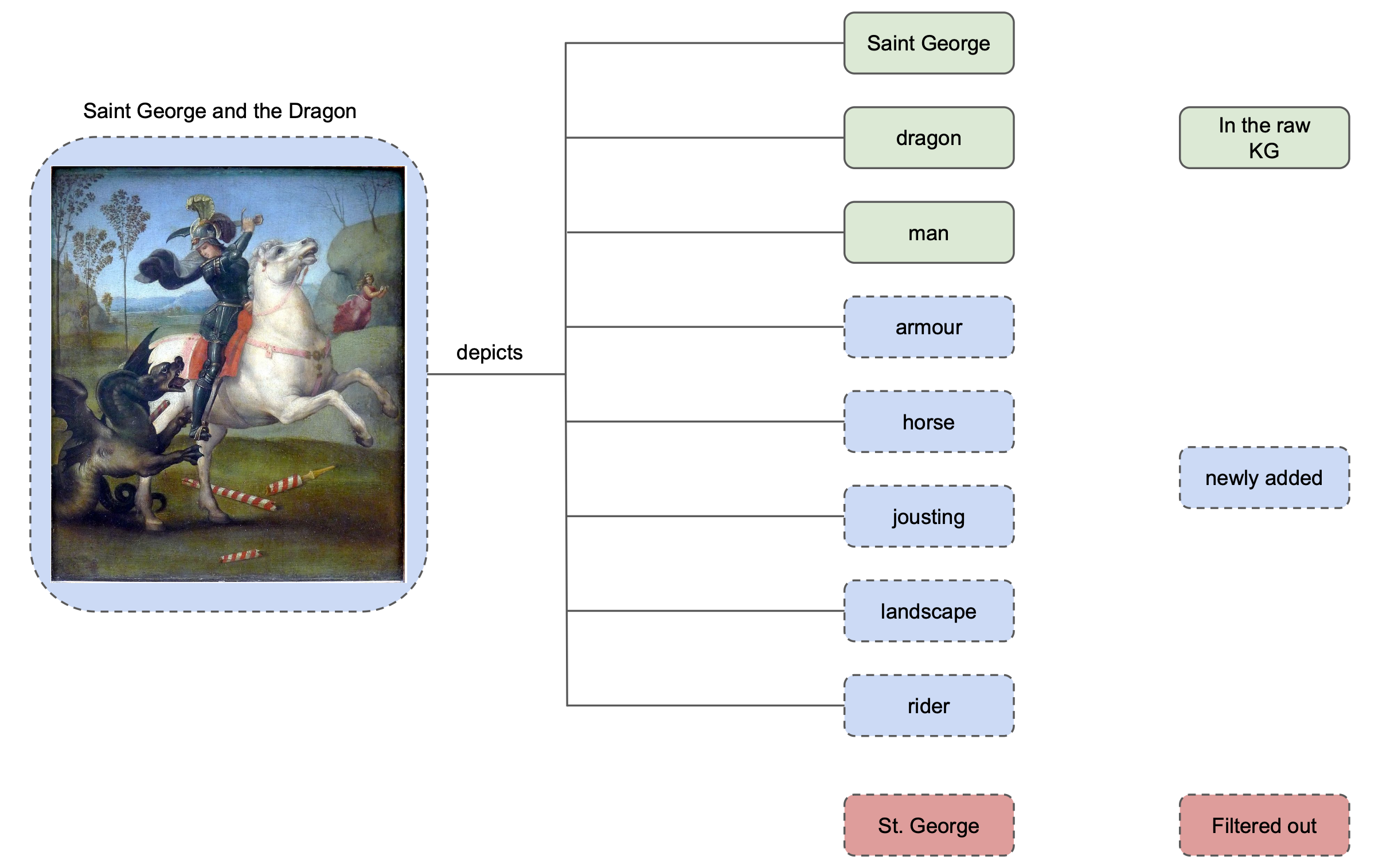}
    \caption{Example of "Saint George and the Dragon": entities in green blocks are already in the dataset, blue and red blocks are newly generated entities, we filter out the red blocks and keep the blue blocks.}
    \label{f1}
\end{figure}

We also present an example from WJocondeMM\_en in Fig. \ref{f1}, where we show "Saint George and the Dragon" as the target painting and "depicts" as the target relation. In the original KG, there are only "Saint George", "dragon" and "man" are associated with this relation, as shown in the green block. After applying the extension pipeline, we added five more entities as shown in the blue block. We can see the added entities like "horse", "jousting", "amour", etc. correctly fit the relation "depicts". Not only are the objects associated with this painting extended, but the scene is also accurately extracted. However, the generated entity "St. George" is considered to be filtered out by our pipeline due to its high similarity with "Saint George," which is already present in our dataset.

\begin{center}
\begin{table}[h]
\caption{Summary of generated entities}
\begin{tabular}{llll}
\hline
 & WJocondeMM\_en& generated & filtered\\
\hline
Number of tail entities & 7,161 & 6,431 & 4,394 \\
Ratio to the raw KG & 100\% & 89.80\% & 61.36\% \\
Ratio of valid entities & - & 100\% & 68.32\% \\
\hline
\end{tabular}
\label{tab:soge}
\end{table}
\end{center}

\begin{center}
\begin{table}[h]
\caption{Results for extended knowledge graph}
\begin{tabular}{lll}
\hline
Element & WJocondeMM\_en& WJocondeMM\_en\_extended\\
\hline
Triples & 19,122 & 23,516 \\
Entities & 7,749 & 12,143\\
Relations & 121& 121\\
\hline
\end{tabular}
\label{tab:stat}
\end{table}
\end{center}

\subsection{Analysis of added entities}

To further evaluate the quality and practical relevance of the proposed knowledge graph extensions, we analyze the semantic nature of the newly added entities by organizing them into a high-level taxonomy. This taxonomy was manually designed to encompass the majority of aspects relevant to cultural heritage objects. This analysis aims to assess whether the extensions capture meaningful descriptive facets useful to end-users and cultural heritage professionals, rather than low-level or redundant visual attributes.

Table~\ref{tab:taxonomy_compare} reports the distribution of entity categories for both the original WJocondeMM\_en dataset and the newly added entities, expressed as percentages of total mentions. In the original dataset, a large proportion of entities fall into the \emph{unknown} category, which mainly includes character names, historical periods, symbolic references, and creation background information that cannot be directly inferred from visual content alone and is therefore primarily derived from textual or archival sources.

In contrast, the added entities are largely distributed across visually grounded and semantically rich categories. Excluding the \emph{unknown} category, the distributions between the original and extended datasets align closely, with categories such as \emph{person}, \emph{nature}, \emph{object}, and \emph{clothing} constituting a substantial portion of the entities in both cases. Contextual categories including \emph{scene}, \emph{building}, and \emph{animal} further enrich the narrative and situational interpretation of artworks. Low-level visual attributes such as \emph{color} and \emph{material} remain comparatively limited, indicating that the proposed pipeline favors conceptually meaningful and user-relevant information.

\begin{center}
\begin{table}[h]
\caption{Comparison of entity taxonomy between the original WJocondeMM dataset and newly added entities (percentage of total mentions)}
\begin{tabular}{lcc}
\hline
Category & WJocondeMM\_en (\%) & Added entities (\%) \\
\hline
Unknown & 55.4 & 1.0 \\
Person & 9.7 & 15.6 \\
Object & 9.2 & 22.5 \\
Nature & 6.3 & 17.5 \\
Clothing & 4.2 & 12.3 \\
Body part & 3.8 & 3.2 \\
Animal & 2.9 & 4.4 \\
Action & 2.8 & 0.8 \\
Building & 2.1 & 7.7 \\
Scene & 1.6 & 6.1 \\
Color & 1.4 & 2.8 \\
Religious & 0.3 & 2.0 \\
Style & 0.2 & 1.7 \\
Material & 0.1 & 2.3 \\
\hline
\end{tabular}
\label{tab:taxonomy_compare}
\end{table}
\end{center}

To further illustrate the semantic nature of the added entities, Table~\ref{tab:taxonomy_examples} provides representative examples for each category. These examples show that the extensions primarily capture interpretable and visually grounded concepts—such as objects, natural elements, architectural structures, human figures, and actions—that are commonly used by art historians and cultural heritage professionals when describing, indexing, or searching visual collections. This qualitative inspection complements the quantitative analysis by demonstrating that the added entities enrich the knowledge graph with semantically meaningful facets that support iconographic interpretation and contextual understanding.

Although our pipeline demonstrates strong performance in introducing new entities, and cross-validation across models is employed to mitigate potential errors and hallucinations, the models may still produce inaccuracies. Gaining a deeper understanding of the conditions under which such errors occur is therefore essential. To this end, we present an analysis of hallucinations in Appendix \ref{hallucination}.

\begin{center}
\begin{table}[h]
\caption{Representative examples of newly added entities by entity taxonomy}
\begin{tabular}{lp{9cm}}
\hline
Category & Example entities \\
\hline
Action & \emph{jousting}, \emph{battle}, \emph{man kneeling}, \emph{horseback riding}, \ldots \\
Animal & \emph{angel}, \emph{horse}, \emph{dog}, \emph{lion}, \emph{cherub}, \ldots \\
Body part & \emph{hand}, \emph{beard}, \emph{hair}, \emph{nude body}, \ldots \\
Building & \emph{arch}, \emph{column}, \emph{castle}, \emph{ceiling}, \emph{wall}, \ldots \\
Clothing & \emph{armor}, \emph{hat}, \emph{dress}, \emph{crown}, \emph{cloak}, \ldots \\
Color & \emph{red dress}, \emph{blue robe}, \emph{dark background}, \emph{shadow}, \ldots \\
Material & \emph{cloth}, \emph{stone}, \emph{gold fabric}, \emph{wooden floor}, \ldots \\
Nature & \emph{sky}, \emph{landscape}, \emph{trees}, \emph{clouds}, \emph{mountains}, \ldots \\
Object & \emph{chair}, \emph{table}, \emph{sword}, \emph{book}, \emph{curtain}, \ldots \\
Person & \emph{man}, \emph{woman}, \emph{child}, \emph{crowd}, \ldots \\
Religious & \emph{cross}, \emph{halo}, \emph{chalice}, \emph{crown of thorns}, \ldots \\
Scene & \emph{cityscape}, \emph{religious scene}, \emph{indoor setting}, \emph{outdoor}, \ldots \\
Style & \emph{portrait}, \emph{still life}, \emph{dramatic lighting}, \ldots \\
\hline
\end{tabular}
\label{tab:taxonomy_examples}
\end{table}
\end{center}

\section{Conclusion and future work}\label{sec6}

This paper contributes to research and applications in the cultural heritage domain by introducing a new knowledge graph focused on French collections. The novelty of this KG is the multi-modal data of text descriptions and visual images for each entity. This dataset can also serve as a new benchmark for KGC, offering an alternative to widely used datasets such as FB15K and WN18. Our dataset is a domain-specific cultural heritage dataset thereby encouraging future research in this area. Finally, we proposed a novel KGC pipeline using state-of-the-art LLMs and vision-language models when external knowledge, such as text and images, is available. 

Our experiments show that, using the proposed pipeline and prompt templates, large models can efficiently extract new entities with a high satisfaction ratio, enabling KGC under the OWA. We also propose a new validation method, applicable within the experimental setup mentioned above. We validated that the models can accurately predict existing entities, demonstrating that this pipeline is both robust and efficient.

For future works, more advanced LLMs and vision-language models can be tested. It would also be interesting to test the pipeline under few-shot learning or more sophisticated fine-tuning scenarios.

\bibliographystyle{ACM-Reference-Format}
\bibliography{acm/sample-base}

\appendix

\section{Performance of embedding models}
\label{Apenx A}
\subsection{Embedding models on WJoconde}

\textbf{Experimental setup}: for the ComplEx\_n3 model, we set $rank=500$, $learning\_rate=5e-2$, $reg=1e-2$. For the TuckER model, we set $lr=5e-4$, $dr=1.0$, $edim=200$, $rdim=200$, $input\_dropout=0.3$, $hidden\_dropout1=0.4$, $hidden\_dropout2 0.5$, $label\_smoothing=0.1$. For other models, we use the Torchkge library and the default parameters proposed by the library. The results are shown in Table~\ref{e1}.

\begin{table}[h]
    \centering
    \caption{Performance of the embedding models on WJconde}
    \begin{tabular}{llll}
    \hline
        models & MR & MRR & Hit@10 \\ \hline
        TransE & 4940 & 0.211 & 0.296 \\ 
        TransH & 5577 & 0.20 & 0.281 \\ 
        TransR & 53222 & 0.005 & 0.010 \\ 
        TransD & 5779 & 0.120 & 0.270 \\ 
        TorusE & 3667 & 0.236 & 0.302 \\ 
        RESCAL & 5035 & 0.215 & 0.270 \\ 
        DistMULT & 4171 & 0.265 & 0.349 \\ 
        HolE & 11900 &  0.148 & 0.210 \\ 
        ComplEx & 2736 &  0.281 & 0.360 \\ 
        ComplEx\_n3 &   & 0.420 & 0.510 \\ 
        TuckER & 453 & 0.540 & 0.655 \\ \hline
    \end{tabular}
    \label{e1}
\end{table}

\subsection{Embedding models on WJoconde\_cleaned}
We use the same setup as before and show the performance of the embedding models in Table~\ref{e2}. This demonstrates that the cleaning process largely improves the quality and reduces the complexity in our dataset.

\begin{table}[h]
    \centering
    \caption{Performance of the embedding models on WJconde\_cleaned}
    \begin{tabular}{llll}
    \hline
        models & MR & MRR & Hit@10 \\ \hline
        TransE & 2919 & 0.197 & 0.303     \\ 
        TransH & 2778 & 0.213 & 0.313 \\ 
        TransR & 34138 & 0.003 & 0.007 \\ 
        TransD & 2188 & 0.155 & 0.343 \\ 
        TorusE & 4517 & 0.282 & 0.362 \\
        RESCAL & 4454 & 0.216 & 0.290 \\ 
        DistMULT & 3514 & 0.275 & 0.350 \\ 
        HolE & 5573   &  0.178 & 0.237 \\ 
        ComplEx & 2039 &  0.273 & 0.365 \\ 
        ComplEx\_n3 &   & 0.603 & 0.515 \\ 
        TuckER & 1743 & 0.603 & 0.706 \\ \hline
    \end{tabular}
    \label{e2}
\end{table}

\subsection{Embedding models on WJocondeMM}

We then report the performance of the embedding models on the WJocondeMM dataset as shown in Table \ref{tab-jocmm}. Here we only tested the graph portion of this dataset, since no text and image are used in embedding models. We use the same setup as before, and the French version of WJocondeMM produces similar results.

\begin{table}[h]
    \centering
    \caption{Performance of the embedding models on WJcondeMM}
    \label{tab-jocmm}
    \begin{tabular}{llll}
    \hline
        models & MR & MRR & Hit@10 \\ \hline
        TransE & 281 & 0.183 & 0.344     \\ 
        TransH & 273 & 0.173 & 0.361 \\ 
        TransR & 1235 & 0.155 & 0.214 \\ 
        TransD & 332 & 0.141 & 0.334 \\ 
        TorusE & 515 & 0.264 & 0.351 \\ 
        RESCAL & 503 & 0.267 & 0.364 \\ 
        DistMULT & 268 & 0.316 & 0.45 \\ 
        HolE & 841 &  0.2 & 0.29 \\ 
        ComplEx & 319 &  0.318 & 0.438 \\ 
        ComplEx\_n3 &   & 0.455 & 0.591 \\ 
        TuckER & 391 & 0.537 & 0.654 \\ \hline
    \end{tabular}
    \label{translation}
\end{table}

\section{Example}
\begin{figure}[h]
    \includegraphics[scale=0.24]{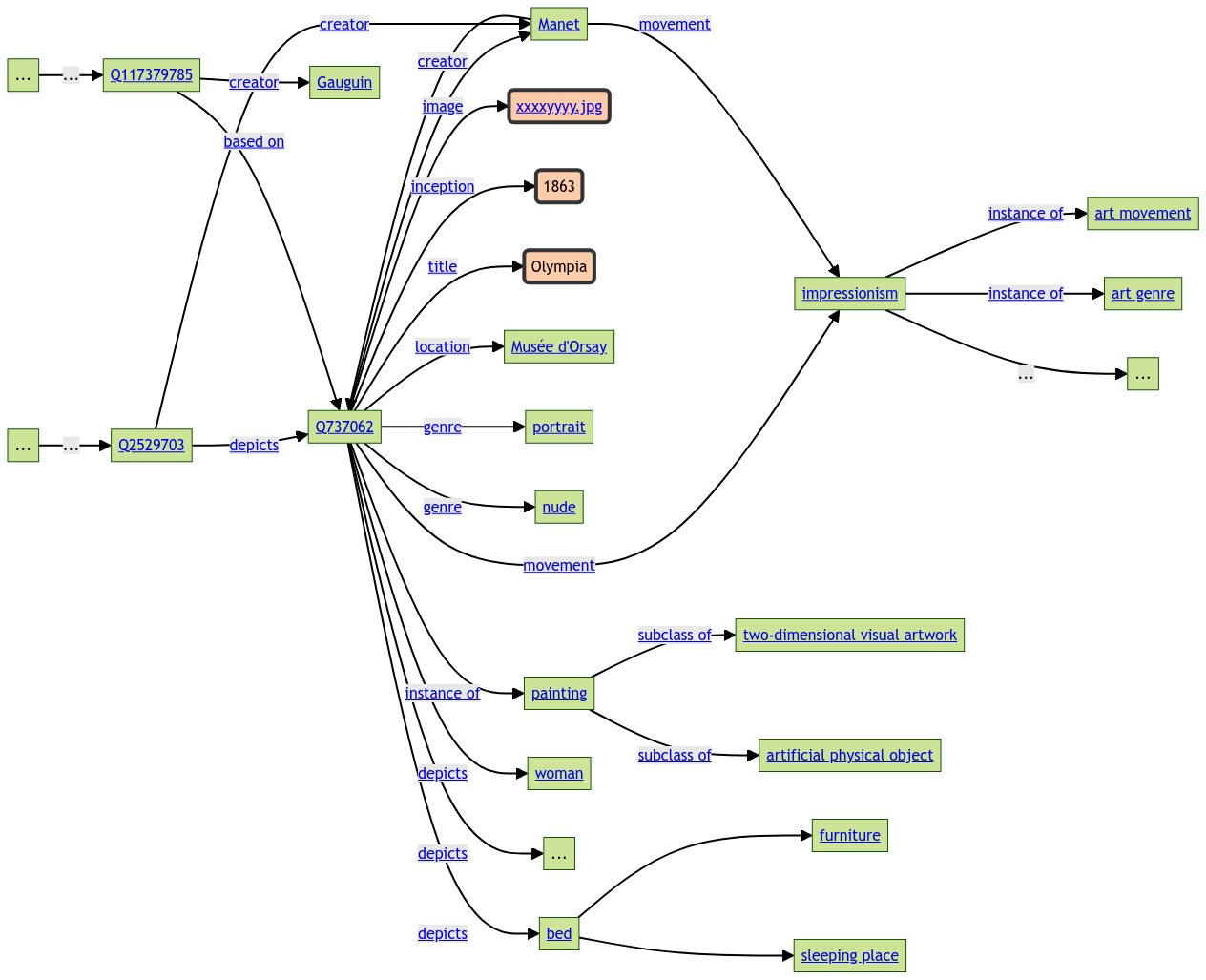}
    \caption{Olympia by Manet, grey denotes properties, green denotes entities, and pink denotes literal values.}
    \label{fig:olympia}
\centering
\end{figure}

Figure ~\ref{fig:olympia} shows a sample. The main entity in this figure is $Q737062$. It has several properties such as \texttt{title} with the value "Olympia" which is a short text, or \texttt{depicts} which links to other entities such as \texttt{bed}. The figure is visible online \footnote{\url{https://givingsense.eu/joconde/Olympia.svg}} in an interactive version.

\section{Hallucination analysis}
\label{hallucination}
In this section, we show some cases where the model fails at predicting reliable entities.

\begin{figure}[h]
    \centering
    \includegraphics[width=0.45\textwidth]{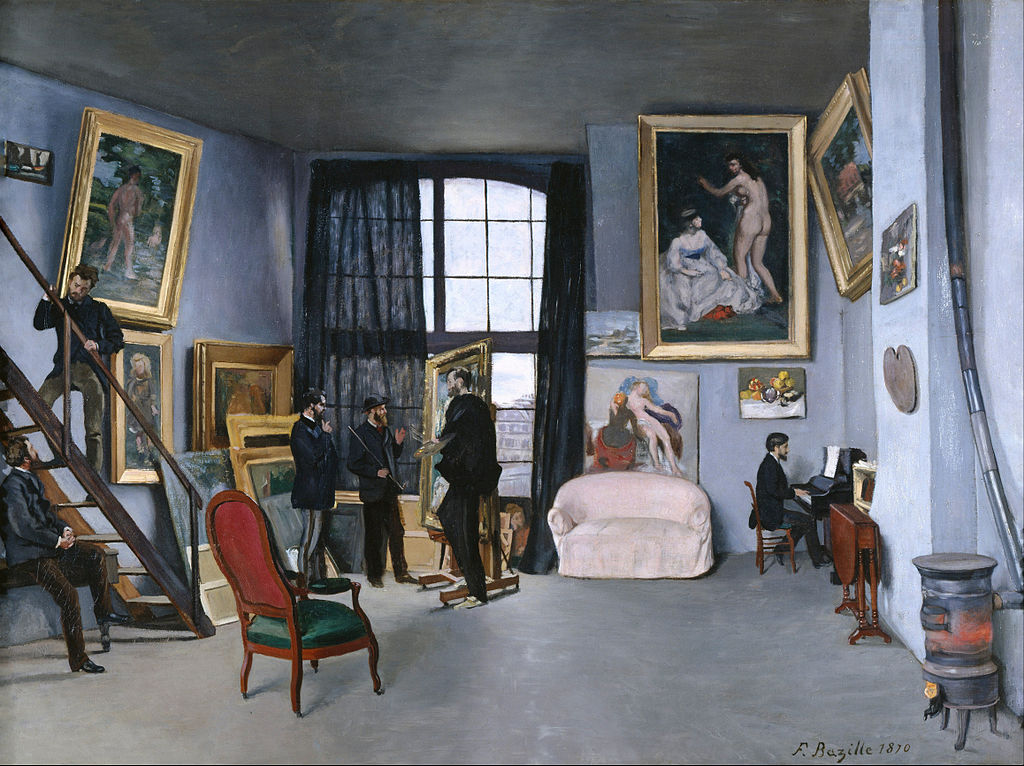}
    \hfill
    \includegraphics[width=0.45\textwidth]{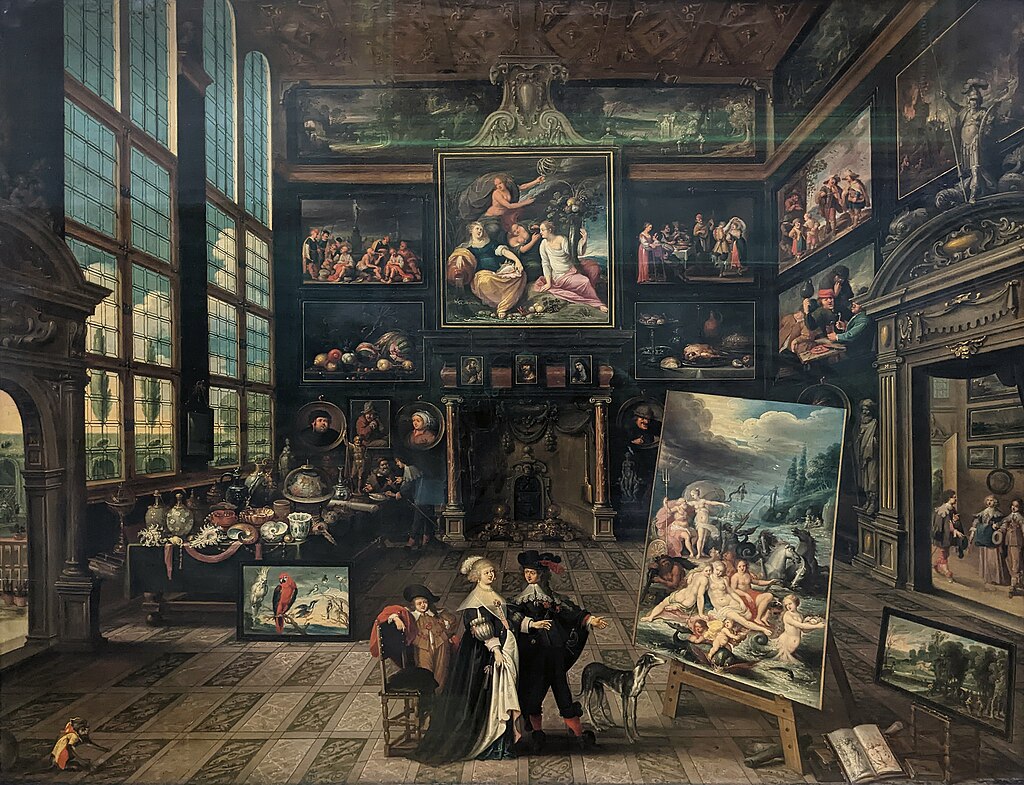}
    \caption{The models predict "candlestick" for both paintings}
    \label{fig:two-images1}
\end{figure}

\begin{figure}[h]
    \centering
    \includegraphics[width=0.45\textwidth]{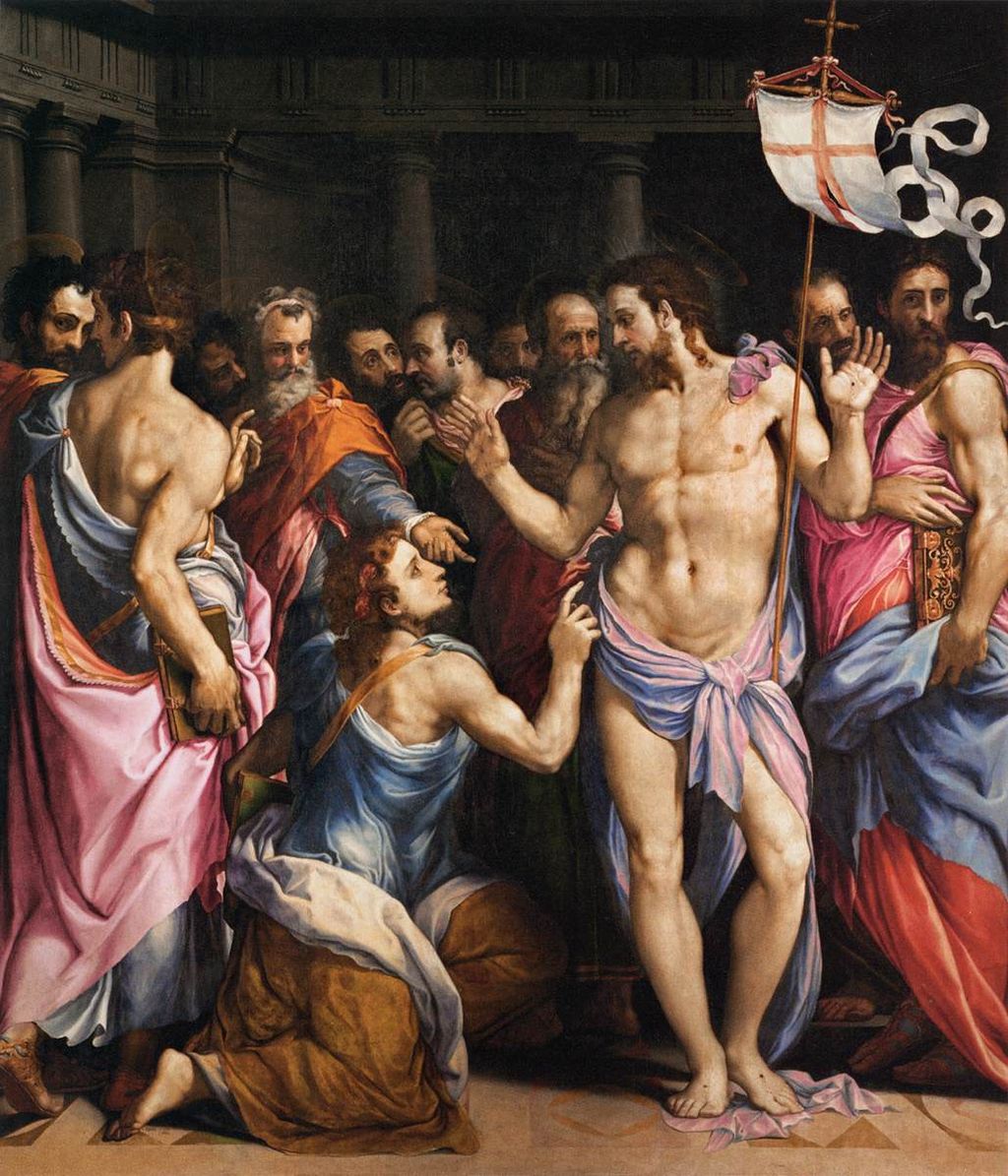}
    \hfill
    \includegraphics[width=0.45\textwidth]{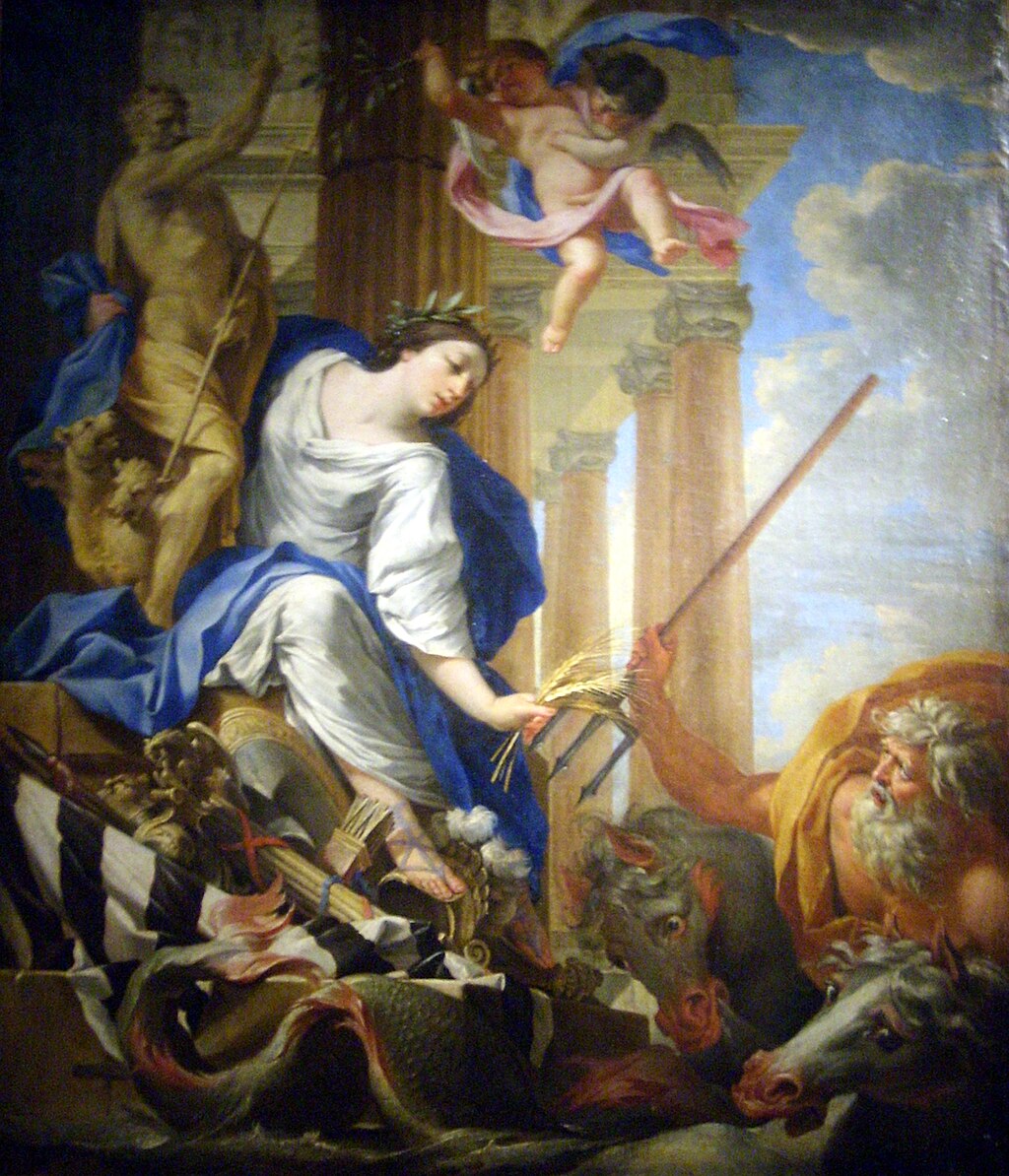}
    \caption{The models predict "shield" for both paintings}
    \label{fig:two-images2}
\end{figure}

During the human evaluation, we identified several recurring errors produced by the models that were not fully addressed by our validation procedures. In particular, the models tend to infer entities based on stereotypical visual cues rather than actual visual evidence. When images exhibit certain stylistic characteristics or contextual elements—such as dim lighting, specific backgrounds, or commonly associated objects—the models often predict entities that typically co-occur in such scenes, even when they are not present. For instance, images depicting a dimly lit interior with tables frequently lead the model to predict the presence of a candlestick, despite its absence as shown in Figure \ref{fig:two-images1}. Similarly, chaotic scenes showing groups of people accompanied by flags or iron objects often trigger predictions of "sword" or "shield" as shown in Figure \ref{fig:two-images2}.

These observations suggest that, in specific visual styles or depicted contexts, the models rely on learned conventions and co-occurrence patterns, generating entities that are commonly associated with those scenarios rather than strictly grounded in the image content. It is worth noting that, in some cases, even human evaluators cannot be fully certain about the presence of such entities when the visual evidence is ambiguous or the image quality is limited. Furthermore, such ambiguous cases are relatively rare in our assessment and will require further analysis.

\newpage
\newpage
\section{Data Availability} \label{data-availability}

The main access for the dataset - KG with texts and images - is:
https://doi.org/10.5281/zenodo.17970221\\

This dataset can also be queried using an interactive SPARQL access point: 

https://datamusee.r2.enst.fr

Select the dataset:

https://datamusee.r2.enst.fr/\#/dataset/WJoconde/query

https://datamusee.r2.enst.fr/\#/dataset/WJoconde\_cleaned/query

https://datamusee.r2.enst.fr/\#/dataset/WJocondeMM\_en/query

https://datamusee.r2.enst.fr/\#/dataset/WJocondeMM\_fr/query





The corresponding access by API:

\hyperlink{https://datamusee.r2.enst.fr/WJoconde/sparql}{https://datamusee.r2.enst.fr/WJoconde/sparql}

\hyperlink{https://datamusee.r2.enst.fr/WJoconde\_cleaned/sparql}{https://datamusee.r2.enst.fr/WJoconde\_cleaned/sparql}

\hyperlink{https://datamusee.r2.enst.fr/WJocondeMM\_en/sparql}{https://datamusee.r2.enst.fr/WJocondeMM\_en/sparql}

\hyperlink{https://datamusee.r2.enst.fr/WJocondeMM\_fr/sparql}{https://datamusee.r2.enst.fr/WJocondeMM\_fr/sparql} \\

A brief description can be found here: 

https://datamusee.wp.imt.fr/2023/12/04/the-joconde-dataset-and-lod/ \\

WJoconde, WJoconde\_cleaned, WJocondeMM\_en and WJocondeMM\_fr each contain csv file with "from" "rel" "to" representing triples.
The file entity2text.txt and rel2text.txt are the corresponding names of all the entities and relations in WJoconde. Moreover, entity2description\_en.csv and entity2description\_fr.csv are the long text descriptions in English and French, respectively. All these files are in the text\_data subdirectory. The names of the images are linked to the URI of the entities (e.g. http//www.wikidata.org/entity/Q100268819
has image httpwwwwikidataorgentityQ100268819.jpg). 

The reference code for the proposed KGC method can be found here: 

https://anonymous.4open.science/r/WJoconde\_KG-49EF




\end{document}